\pdfoutput=1
\documentclass[11pt]{article}

\usepackage[final]{acl}

\usepackage{times}
\usepackage{latexsym}
\usepackage[T1]{fontenc}
\usepackage[utf8]{inputenc}
\usepackage{microtype}
\usepackage{inconsolata}
\usepackage{booktabs}
\usepackage{amsmath}
\usepackage{amssymb}
\usepackage{enumitem}
\usepackage{graphicx}
\usepackage{float}
\usepackage{multirow}

\title{Detection $\neq$ Reliable Control: Decodable Empathy Directions Yield \\ at Most Partial Shifts in Automated Empathy Scores}

\author{Haoran Jisun \\
  University of Southern California \\
  \texttt{hjisun@usc.edu}}

\begin{document}
\maketitle

\begin{abstract}
A decodable ``empathy'' direction is routinely read as a causal lever, conflating decodability, automated-metric control, and human-perceived change. We test this for two EPITOME-derived facets---\emph{Recognition} (cognitive) and \emph{Resonance} (affective)---in three instruction-tuned LLMs, scoring every intervention with two LLM judges and a discriminative EPITOME classifier, each gated by an emotional-vs-neutral positive control. The control passes for the affective facet across all automated instruments, but cognitive range is inconsistent across them. Both facets remain decodable after residualizing against a sentence-embedding-derived surface score, and steering can substantially rewrite the text. Yet \emph{adding} the Resonance direction raises the affective score only partially---in Qwen by $+0.29$ ($\approx\!26\%$ of the natural gap). A direct between-direction contrast confirms the shift is facet-specific in Qwen and Llama (not Gemma); we do not, however, establish a matching human-perceived change. Additive cognitive steering produces no measurable change, but a within-domain control shows the cognitive instrument is too coarse to resolve the differences such steering would produce---\emph{unmeasurable}, not a clean null. By contrast, Gemma Recognition ablation lowers the classifier's cognitive score even after adjusting for response length. Detection does not imply reliable control under global interventions, and cognitive-empathy claims warrant an explicit measurement-sensitivity check.
\end{abstract}

\section{Introduction}

Affective concepts are linearly decodable from LLM hidden states, and some are causally steerable \citep{tigges2024sentiment, turner2023actadd, li2023iti, chen2025persona}. But decodability is a weak basis for causal claims: high probe accuracy need not mean that a property is genuinely encoded \citep{hewitt2019control}, or that the model \emph{uses} it \citep{elazar2021amnesic}, steering vectors are brittle and unreliable \citep{tan2024steering, canby2025probing}, and apparently meaningful directions can dissolve under control---an \emph{interpretability illusion} affecting directions no less than individual neurons \citep{bolukbasi2021illusion}.

Empathy is a high-stakes place to test this: LLMs are increasingly deployed in emotional-support settings. Psychology has long treated empathy as multidimensional, separating a \emph{cognitive} component---inferring another's state through perspective-taking---from an \emph{affective} component---resonating with their feelings \citep{davis1983empathy, jolliffe2006empathy}. In LLM \emph{behavior}, the same two-factor structure appears in some models but not others---reproduced by GPT-4, not by Llama-3 \citep{YU2025100233}---and whether it corresponds to distinct \emph{representations} an intervention can target is unknown; whether detection implies control may itself be facet-dependent. We operationalize the two facets via EPITOME, which decomposes textual empathy into \emph{Emotional Reactions} (affective resonance), \emph{Interpretations} (cognitive recognition), and \emph{Explorations} \citep{sharma2020epitome}. Throughout, our \emph{Recognition} facet denotes this \emph{interpretive} cognitive empathy---communicating an inferred understanding of the seeker's experience---not emotion recognition or affect classification. All claims here accordingly concern \emph{text-expressed} empathy---empathic communication in EPITOME's sense---not internal experience; ``empathy'' below is shorthand for this operationalization. Our central contribution is to keep three usually-merged questions apart---decodability, automated-metric control under a global intervention, and human-perceived change---each interpretable only once a positive control establishes the instrument's sensitivity. We probe Llama-3.1-8B \citep{llama3herd}, Gemma-2-9b \citep{gemma2}, and Qwen2.5-7B \citep{qwen25}, and---because empathy is difficult to operationalize and judge reliably---read every outcome through two LLM judges and a discriminative EPITOME classifier, supplemented by a human calibration check, all gated by a positive control. We find:
\begin{itemize}[noitemsep,topsep=2pt,leftmargin=*]
    \item \textbf{Detectable.} EPITOME-derived Recognition and Resonance directions are linearly decodable and remain so after residualizing against a sentence-embedding-derived surface score.
    \item \textbf{A measurement caveat (the central measurement finding).} A positive control---separating emotional from neutral prompts---shows the \emph{affective} facet has robust range across all automated instruments (all $d{\ge}2.6$; inter-judge $\alpha{=}0.70$--$0.84$), whereas cognitive range is instrument-dependent: GPT-4o separates emotional from neutral in all three models, the second judge only in Gemma, and the classifier only in Llama and Gemma ($d{=}0.62/0.67$), not Qwen ($0.00$); the judges agree only weakly on it ($\alpha\le 0.44$). A finer, warmth-matched \emph{within-domain} control shows that even this classifier cannot resolve differences at the scale on which an additive-steering effect would have to register. We scope every claim to the facet, model, and resolution a given instrument can measure.
    \item \textbf{Affective: a graded, sub-maximal lever on the automated metric---not a clean null, not global control.} \emph{Adding} the Resonance direction raises the classifier-measured affective score: in Qwen the $\alpha{=}0{\to}{+}8$ endpoint rises $+0.29$ ($p{<}10^{-4}$; ${\approx}26\%$ of the natural gap, full span ${\approx}39\%$). Separately, a repeated-measures trend test confirms an ordered dose--response---our primary result. The effect is present but \emph{exploratory} in Llama (only at the raw, surface-entangled peak) and practically negligible in Gemma. It is facet-differentiated in Qwen and Llama---a direct paired contrast confirms Resonance, not Recognition, raises the classifier's affective score (null in Gemma)---but shown on classifiers, not human readers, and nowhere approaches the gap.
    \item \textbf{Cognitive: a length-controlled necessity signal in one model; additive steering unmeasurable.} All-layer ablation lowers the classifier-measured cognitive score only for Gemma Recognition. A within-prompt length control dissociates the two instruments: the LLM-judge drop is largely length-mediated, but the discriminative classifier's drop survives adjustment for length---its bi-encoder truncates at $64$ tokens, and the drop is not attributable to reduced elaboration alone---while affective scores stay at ceiling (Appendix~\ref{app:length}). Additive Recognition steering is flat in every model, but a within-domain control shows the cognitive instrument lacks demonstrated sensitivity at the scale on which an additive-steering effect would have to register---so the additive cognitive null is \emph{unmeasurable}, not established.
\end{itemize}
\section{Method}

\paragraph{Probes.} For each facet we sample $100$ high (Level~2) and $100$ low (Level~0) EPITOME items, format them as a \texttt{Client}/\texttt{Counselor} exchange read under a counselor system prompt, and extract the last-token, pre-generation residual-stream activation at every layer. The probe direction is the difference of class means; per-layer effect size is Cohen's $d$.

\paragraph{Surface-text residualization.} High decodability can reflect lexical surface form rather than an internal state \citep{lee2023context}. We compress a sentence embedding (MiniLM\,/\,MPNet; \citealp{wang2020minilm, song2020mpnet, reimers2019sentencebert}) to a scalar surface score via a cross-validated classifier, regress the scalar probe projection on it, and recompute $d$ on the residuals, selecting each facet's peak layer by $\arg\max$ residualized $d$. This residualization is a \emph{detection control and a layer-selection criterion}: it shows separation is not driven by the surface scalar, and it picks the peak layer. The steering vector itself is the raw class-mean difference at that layer, \emph{not} a residualized vector; we therefore claim a residualized \emph{detection} result, not a deconfounded intervention direction, and report held-out cross-validated $d$ for both forms (Table~\ref{tab:main}).

\paragraph{Causal steering.} At the peak layer we add $\alpha\,\sigma\,\hat{\mathbf{d}}$ (magnitude in projection-SD units, $\alpha\in[-8,8]$) and generate to $50$ prompts ($40$ emotional, drawn from an emotional-support corpus \citep{liu2021esconv}, and $10$ neutral, to retain rating headroom and diagnose ceilings). Generation is greedy and deterministic (\texttt{do\_sample}=False, up to $90$ new tokens for steering; ablation and direction-estimation caps differ, see Appendix~\ref{app:repro}): each $(\text{prompt},\alpha)$ condition yields one completion, so the sweep carries no within-condition sampling noise and each per-prompt slope reflects steering alone. The GPT-4o judge scores at temperature $0$, the second judge (\texttt{moonshot-v1-128k}) at temperature $1$. Because each direction is scaled by its \emph{own} projection SD $\sigma$, the empathy directions receive a larger raw-norm perturbation than a random direction at the same layer; we control for this asymmetry with the random-direction floor below. A blind GPT-4o judge rates cognitive and affective empathy on independent $1$--$7$ scales, in shuffled order. We require a concept-specific causal effect to move the \emph{matched} dimension more than the other dimension \emph{and} more than a random-direction floor, without degeneration; we report bootstrap $95\%$ CIs on the matched effect and Bonferroni-correct lexical analyses. We treat a one-point shift on the $1$--$7$ scale as the threshold of practical significance; under the paired ablation design the minimum detectable standardized effect is $d_z{\approx}0.40$, i.e.\ $\approx 0.16$ points at the within-prompt difference SD $\approx 0.39$ (Appendix~\ref{app:stats}). We steer at both the residualized peak (principled) and the raw Cohen's-$d$ peak (the surface-entangled layer). Addition tests \emph{sufficiency} (limited by the rating ceiling on emotional prompts) and ablation tests \emph{necessity} (with room to register a drop); we read necessity primarily off ablation. \emph{Analysis hierarchy:} our \emph{primary} analysis is Qwen Resonance at the residualized-selected layer (we call it primary, not confirmatory: it was not preregistered); all other analyses are secondary or exploratory. Implementation details and random seeds are in Appendix~\ref{app:repro}.

\paragraph{Production-side (response-token) direction.} A reading-time probe direction need not coincide with the response-token direction persona vectors are built from \citep{chen2025persona}. We therefore estimate a second direction as the difference of class means over \emph{response-token} activations of completions generated from high- vs.\ low-Recognition EPITOME inputs, apply the same surface-text residualization, and (i)~measure its cosine to the read-time probe direction and (ii)~steer with it at its own peak layer.

\paragraph{Directional ablation.} As a more extensive intervention than additive steering, we project the \emph{single} peak-layer unit direction out of the residual-stream output of \emph{every} layer at \emph{every} token position during generation (the residual stream is one shared space across layers, so the same direction is removed throughout)---the all-layer, all-position ablation \citep[cf.][]{arditi2024refusal} used to establish a causal sentiment direction in \citet{tigges2024sentiment}---and compare judged empathy to a floor of five random-direction ablations. The floor aggregates the five random directions per prompt, so the matched-vs-floor comparison is paired by prompt; we test it with a sign-flip permutation test (and Wilcoxon signed-rank), not an unpaired statistic.

\paragraph{Output-change diagnostic.} To confirm the intervention is potent---i.e.\ that a null on empathy is not a null intervention---we measure, per condition, the MPNet cosine between the $\alpha{=}0$ and $\alpha{=}{+}8$ generations (lower $=$ larger text change), benchmarked against (a) a random-direction floor (matched in projection-SD units, hence smaller in raw norm; see \emph{Causal steering}) and (b) a cross-prompt floor (cosine between unrelated $\alpha{=}0$ generations). We flag degeneration by the unique-token ratio (collapse at ${<}0.5$); where a principled peak degenerates at $|\alpha|{=}8$, we read the diagnostic at the largest non-degenerate magnitude.

\paragraph{Additional instruments and a built-in positive control.} Because LLM-as-a-judge ratings carry systematic biases \citep{zheng2023judging, li2026counselbench} and empathy is hard to judge reliably, we evaluate every steered and ablated generation with two further automated instruments alongside the primary GPT-4o judge---(i)~a \emph{second} LLM judge (\texttt{moonshot-v1-128k}) under a parallel $1$--$7$ cognitive/affective rubric (not identical in wording; Appendix~\ref{app:repro}), yielding Krippendorff's $\alpha$ \citep{krippendorff2011computing} per facet, and (ii)~a \emph{discriminative, non-LLM} EPITOME classifier---the \citet{sharma2020epitome} bi-encoder, empathy-identification head only (rationale head dropped), re-trained on the EPITOME Reddit corpus---scoring each (prompt, response) pair to an expected level in $[0,2]$ (held-out macro-F1 and per-level confusion in Appendix~\ref{app:clf})---plus (iii)~a blind six-rater \emph{pairwise-comparison} panel: $16$ hand-built calibration pairs and $8$ Qwen Resonance $\alpha{=}0$-vs-${+}8$ output pairs as a steered-output spot-check (Appendix~\ref{app:rater}). Crucially, before interpreting any null we run a \emph{positive control}: does each instrument separate responses to \emph{emotional} from \emph{neutral} prompts? This is a coarse range check: passing it does not by itself establish sensitivity to the smaller, within-domain changes steering induces. We therefore additionally run a finer, \emph{within-emotional-domain} control (Appendix~\ref{app:within}): $16$ warmth-matched pairs differing only in cognitive specificity, which a valid cognitive instrument should separate while a warmth-matched affective instrument should not. Because $10$ neutral prompts cannot power a cognitive control (minimum detectable $d\approx 1.0$), we expand the control's neutral set to $n{=}40$ with $30$ additional Dolly factual prompts ($\alpha{=}0$ baselines only; the steering set is unchanged), giving $40$ emotional vs.\ $40$ neutral per model; under expansion the cognitive effect sizes hold ($d$ $0.70/0.67 \to 0.62/0.67$ for Llama/Gemma) while becoming significant, and we report the $n{=}40$ values. An instrument that fails this control on a facet cannot adjudicate a steering null there, and we say so rather than reporting the null. The gating is asymmetric by design: an effect that registers demonstrates the instrument's sensitivity, whereas a flat curve does not---so demonstrated resolution gates a null, not a positive effect. Because the classifier reads (prompt, response) pairs, this contrast could reflect detection of the emotional \emph{prompt} rather than empathy in the \emph{response}; we re-score the same control with the seeker tower neutralized (constant placeholder, so only the response varies), which leaves the affective separation essentially unchanged ($d{=}3.6/3.1/4.5$ response-only vs.\ $3.5/3.1/4.8$ with the prompt, for Llama/Qwen/Gemma; stable across two placeholders): the separation is carried by the response text.

\section{Results}

\paragraph{Both facets are decodable and survive residualization.} Recognition peaks in mid-to-late layers, Resonance earlier; residualized effect sizes remain substantial in every model (Table~\ref{tab:main}; Llama Resonance the smallest and most fold-variable). Geometry between the two directions is model-dependent (near-orthogonal in Llama to substantially anti-aligned in Gemma), so we treat it as descriptive only.

\begin{figure*}[t]
\centering
\includegraphics[width=0.92\textwidth]{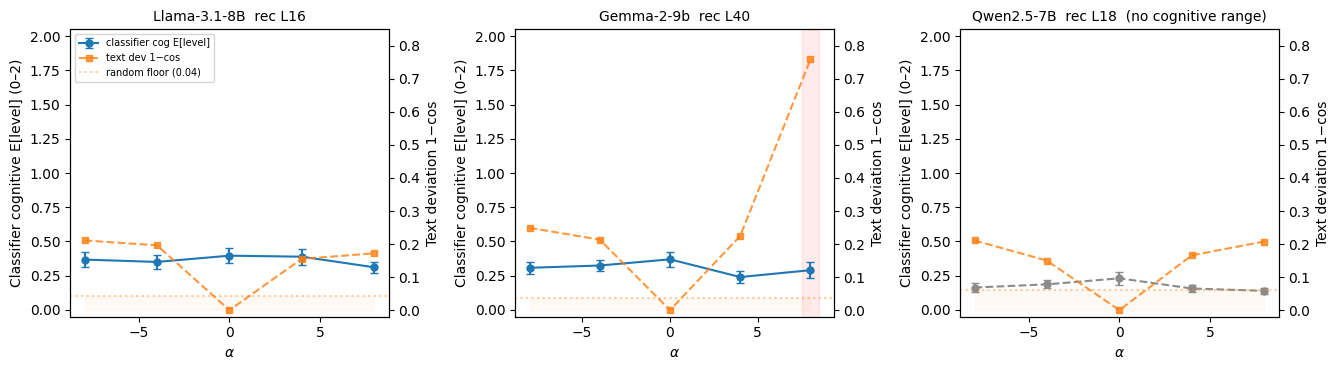}
\caption{\textbf{Recognition steering changes text without a measurable cognitive shift.} Recognition direction added at each model's residualized peak across $\alpha\in[-8,8]$. Blue (left axis): classifier cognitive score ($E[\text{level}]$, $0$--$2$) stays flat for Llama and Gemma; Qwen is greyed (no cognitive range, $d{=}0.00$). Even where the classifier has coarse range, a within-domain control shows it cannot resolve within-domain cognitive differences (Appendix~\ref{app:within}), so these curves leave additive cognitive steering \emph{unmeasurable}. Orange, dashed (right axis): text deviation $1{-}\cos(\alpha{=}0,\alpha)$ rises with $|\alpha|$, confirming a potent intervention. Gemma at $\alpha{=}{+}8$ is degenerate (shaded); its potency is read at $\alpha{=}{+}4$, and the primary-judge curve is likewise flat.}
\label{fig:steering}
\end{figure*}

\paragraph{Steering moves the text; affect moves partway, cognition unmeasurable.} A random direction matched in projection-SD units barely perturbs the output ($\alpha{=}0$ vs.\ $\alpha{=}{+}8$ cosine $0.94$--$0.98$ across models), whereas the empathy directions drive it substantially lower---to $0.78$--$0.83$ at the principled recognition peaks---while remaining coherent (unique-token ratio $\approx 0.8$; Figure~\ref{fig:steering}). The intervention is thus potent on the text; its effect on \emph{empathy}, however, is bounded by which facet---and at what resolution---we can measure (Appendix~\ref{app:example}).

\begin{figure*}[t]
\centering
\includegraphics[width=0.92\textwidth]{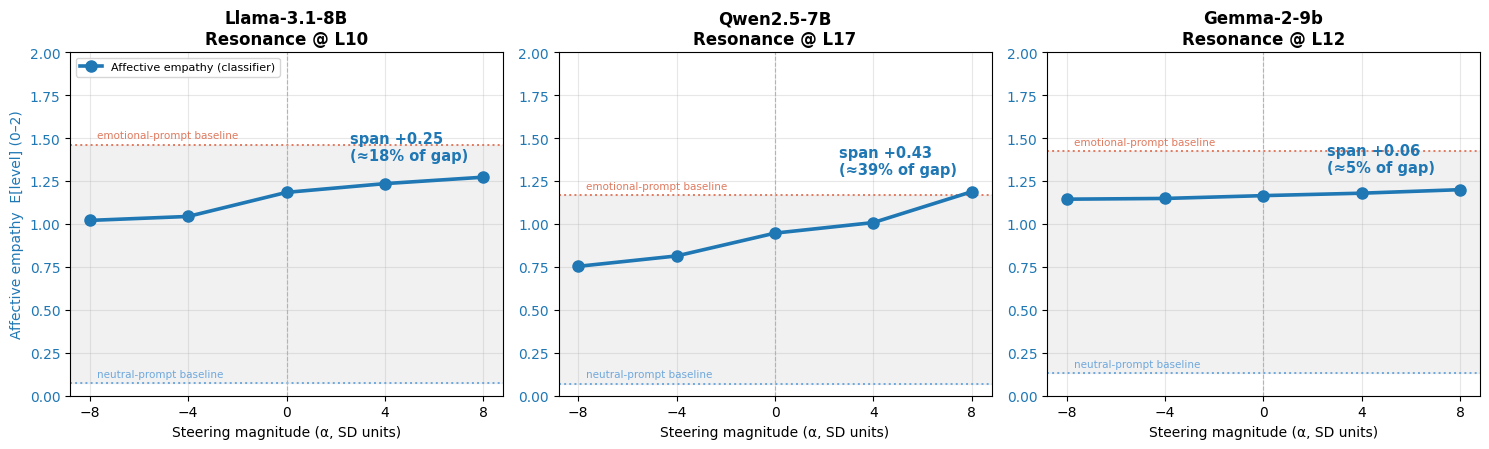}
\caption{\textbf{Resonance steering partially shifts the automated affective metric.} Resonance steering at the resonance peak, scored by the discriminative EPITOME classifier (affective $E[\text{level}]$, $0$--$2$), across $\alpha\in[-8,8]$. The shaded band spans each model's natural neutral-prompt (lower) to emotional-prompt (upper) baseline; the sweep raises the classifier-measured affective score along a graded dose--response---primary in Qwen ($+0.29$ positive gain $\alpha{=}0{\to}{+}8$, $\approx 26\%$ of that gap; $\approx 39\%$ over the full span), exploratory in Llama (raw, surface-entangled peak L10), practically negligible in Gemma; Page's trend test on the matched-prompt design gives $p{=}10^{-4}$ (the floor of $10^4$ permutations), $p{=}10^{-4}$, and $p{=}0.0016$ respectively (text). This is the one setting where a global additive intervention measurably---if sub-maximally---shifts the automated affective metric; contrast the flat cognitive curves of Figure~\ref{fig:steering}.}
\label{fig:affective}
\end{figure*}

On the \emph{affective} facet---where every automated instrument has range---\emph{adding} the Resonance direction raises the classifier-measured affective score ($0$--$2$; Figure~\ref{fig:affective}) by a graded, sub-maximal amount that falls off sharply across models. From baseline to $\alpha{=}{+}8$ the gain is $+0.29$ (Qwen), $+0.11$ (Llama), and $+0.05$ (Gemma); a paired $\alpha{=}0$ vs.\ ${+}8$ test over the same prompts puts each above baseline (one-sided Wilcoxon $p{<}10^{-4}$, $p{=}0.001$, $p{=}0.006$; bootstrap $95\%$ CIs all excluding zero). Against each model's $n{=}40$ emotional-vs-neutral gap---$1.10$ (Qwen), $1.39$ (Llama), $1.29$ (Gemma)---this is $\approx 26\%$/$8\%$/$4\%$ respectively; the full $-8\!\to\!+8$ span---dose--response evidence, not the gain from \emph{adding}---reaches $+0.43$/$+0.25$/$+0.06$ ($\approx 39\%$/$18\%$/$5\%$). Llama's gain, though significant, appears only at the raw, surface-entangled peak L10 (it vanishes at the residualized L5), so we keep it \emph{exploratory}; Gemma's is practically negligible.

The dose--response is robust, not an averaging artifact. Because the same $50$ prompts appear at every $\alpha$, Page's trend test \citep{page1963ordered} (one-sided; $10^4$ within-prompt permutations) gives $p{=}10^{-4}$ (Qwen), $10^{-4}$ (Llama), and $0.0016$ (Gemma), with prompt-clustered OLS slopes $+0.027$/$+0.017$/$+0.0035$ (Gemma's CI $[-0.0004,+0.0076]$ touches zero). In a joint Benjamini--Hochberg \citep{benjamini1995controlling} family of $54$ contrasts---$36$ steered-vs-baseline ($\alpha\in\{\pm4,\pm8\}$; paired Wilcoxon) plus $18$ ablation---$7$ steering contrasts survive ($q_{\min}{<}0.001$), all Resonance (four Qwen, three Llama), with no Recognition contrast surviving. Each tested Qwen magnitude differs from baseline after correction, separate from the ordered Page trend; no Gemma steering magnitude survives. The effect lives in the emotional prompts (their slope exceeds the pooled one; neutral prompts contribute $\approx 0$), and the GPT-4o judge corroborates Qwen (full-span $+0.66$ on $1$--$7$, $p{=}0.003$) while the second does not.

The rise is facet-differentiated, not generic score inflation. A direct paired contrast (Appendix~\ref{app:crossfacet}) shows Resonance raises ER more than Recognition does in Qwen ($+0.31$, $95\%$ CI $[0.21,0.43]$, $p{<}10^{-4}$) and Llama ($+0.14$, $[0.06,0.23]$, $p{<}10^{-3}$) but not Gemma ($-0.01$, $p{=}0.83$); we treat this between-direction contrast as the primary specificity test. A cross-facet matrix corroborates it: \emph{adding Recognition never raises the affective score} (Recognition$\to$ER n.s.\ in all three), while adding Resonance raises ER and, in Qwen and Llama, \emph{lowers} IP, ruling out a uniform lift of all EPITOME features. As complementary evidence, the $+0.29$/$+0.11$ Resonance shift also exceeds all $12$ norm- and layer-matched random directions (floor mean $+0.016$/$-0.006$; ${\approx}5.0$/$2.9$ SDs above it for Qwen/Llama); with $12$ controls the smallest attainable empirical one-sided $p$ is $1/13{=}0.077$, so we read the floor as descriptive support, not a standalone test. The metric is, however, an EPITOME-ER classifier reading directions built from EPITOME-ER labels, and an ER rise driven by added emotional \emph{register} cannot be separated from one driven by added \emph{empathy} without human evaluation (Limitations).

The model-level pattern in text change tracks this gradient. The Resonance direction substantially rewrites the output in Qwen ($\alpha{=}0$ vs.\ ${+}8$ cosine $0.78$ vs.\ a random floor $0.94$) and at Llama's raw peak L10 ($0.83$ vs.\ $0.96$), but barely moves it at Gemma's L12 or Llama's residualized L5 (both $0.90$, near the random floor): where the text moves substantially, affect moves; where it barely moves, the affective gain is absent or practically negligible. At the residualized L5 the intervention barely changes the output and the judge registers nothing. Gemma's flatness is not under-dosing: escalating to $\alpha\in\{12,\dots,24\}$ at L12 drives text change to the Recognition direction's raw-peak level without degeneration, yet the affective score merely saturates (peak ${+}0.04$ at $\alpha{=}16$, back to ${+}0.01$ by $\alpha{=}24$; no cognitive pattern, $|\Delta_{\text{cog}}|\le 0.06$). The affective direction is thus at most a partial, graded lever---primary in Qwen, exploratory in Llama, practically negligible in Gemma---never reliable control.

On the \emph{cognitive} facet the recognition-steering curves are flat in every model and on every instrument (Figure~\ref{fig:steering}), but a flat curve is only as informative as the measure's sensitivity. For the discriminative classifier of Figure~\ref{fig:steering}, the emotional-vs-neutral control establishes coarse cognitive range only for Llama and Gemma, not Qwen. To test the resolution that actually matters here, we run the within-domain control (Appendix~\ref{app:within}): on $16$ warmth-matched pairs differing only in cognitive specificity (warmth matched, ER $E[\text{level}]$ gap $-0.09$), the cognitive classifier does \emph{not} reliably separate the high- from the low-understanding member ($d_z{=}0.38$; Wilcoxon $p{=}0.16$, n.s.), resolving only the very starkest hand-built contrasts. The cognitive instrument therefore lacks demonstrated sensitivity to fine within-domain cognitive differences, so for \emph{all three} models the flat additive curve cannot distinguish no effect from an effect the instrument cannot resolve: additive cognitive steering is \emph{unmeasurable}, not a clean null. (Lexical analyses are likewise null---Llama $0/128$ effects survive Bonferroni---and large $|\alpha|$ induces degeneration rather than empathy; Gemma's principled recognition peak (L40) collapses at $\alpha{=}{+}8$, unique-token ratio $0.13$, cosine $0.24$, so we read its potency at $\alpha{=}{+}4$, cosine $0.78$, coherent.) Additive control is thus, where measurable at all, a graded sub-maximal \emph{affective} lever (largest in Qwen); on the cognitive facet, additive steering is unmeasurable.

\begin{table*}[t]
\centering
\small
\begin{tabular}{@{}llcccc@{}}
\toprule
\textbf{Model} & \textbf{Facet (dim.)} & \textbf{Peak} & \textbf{Resid.\ $d$} & \textbf{Steer $\Delta$ ($\alpha{:}{-}8{\to}{+}8$)} & \textbf{$\cos$ emp/rand} \\
 & & (layer) & & (classifier) & ($\alpha{=}0$ vs ${+}8$) \\
\midrule
\multirow{2}{*}{Llama-3.1-8B} & Recognition (cog) & L16 & $0.99{\pm}0.38$ & $\approx 0$ (n.m.) & $0.83 / 0.96$ \\
 & Resonance (aff) & L5$^{\ddagger}$ & $0.67{\pm}0.47$ & $+0.11$ ($8\%$; span $+0.25$) & --- \\
\midrule
\multirow{2}{*}{Gemma-2-9b} & Recognition (cog) & L40 & $1.10{\pm}0.53$ & $\approx 0$ (n.m.) & $0.78^{\dagger} / 0.98$ \\
 & Resonance (aff) & L12 & $0.86{\pm}0.59$ & $+0.05$ ($4\%$; span $+0.06$) & --- \\
\midrule
\multirow{2}{*}{Qwen2.5-7B} & Recognition (cog) & L18 & $1.08{\pm}0.27$ & $\approx 0$ (n.m.) & $0.79 / 0.94$ \\
 & Resonance (aff) & L17 & $1.02{\pm}0.60$ & $+0.29$ ($26\%$; span $+0.43$) & --- \\
\bottomrule
\end{tabular}
\caption{\textbf{Detection and additive-steering results.} \textbf{Resid.\ $d$}: mean $\pm$ SD of the held-out residualized Cohen's $d$ across five nested folds (peak layer and surface regression fit on training folds, evaluated held-out); all six raw directions cross-validate to held-out $d{>}1.3$ ($1.33$--$2.32$; one-sided permutation $p{=}0.001$ each). \textbf{Steer $\Delta$}: matched-head EPITOME-classifier change from $\alpha{=}0$ to ${+}8$, with the full $-8{\to}{+}8$ span in parentheses; ``n.m.''\ ${=}$ unmeasurable given insufficient within-domain cognitive sensitivity (Appendix~\ref{app:within}). The Resonance gain is $\approx 26\%$ of Qwen's natural gap ($\approx 39\%$ over the span). \textbf{$\cos$ emp/rand}: $\alpha{=}0$ vs.\ ${+}8$ MPNet cosine for the empathy vs.\ a projection-SD-matched random direction (lower ${=}$ larger text change). $^{\dagger}$Gemma read at $\alpha{=}{+}4$ ($+8$ degenerates). $^{\ddagger}$Llama Resonance: detection peak L5, steering and span read at raw peak L10.}
\label{tab:main}
\end{table*}

\begin{figure*}[t]
\centering
\includegraphics[width=0.88\textwidth]{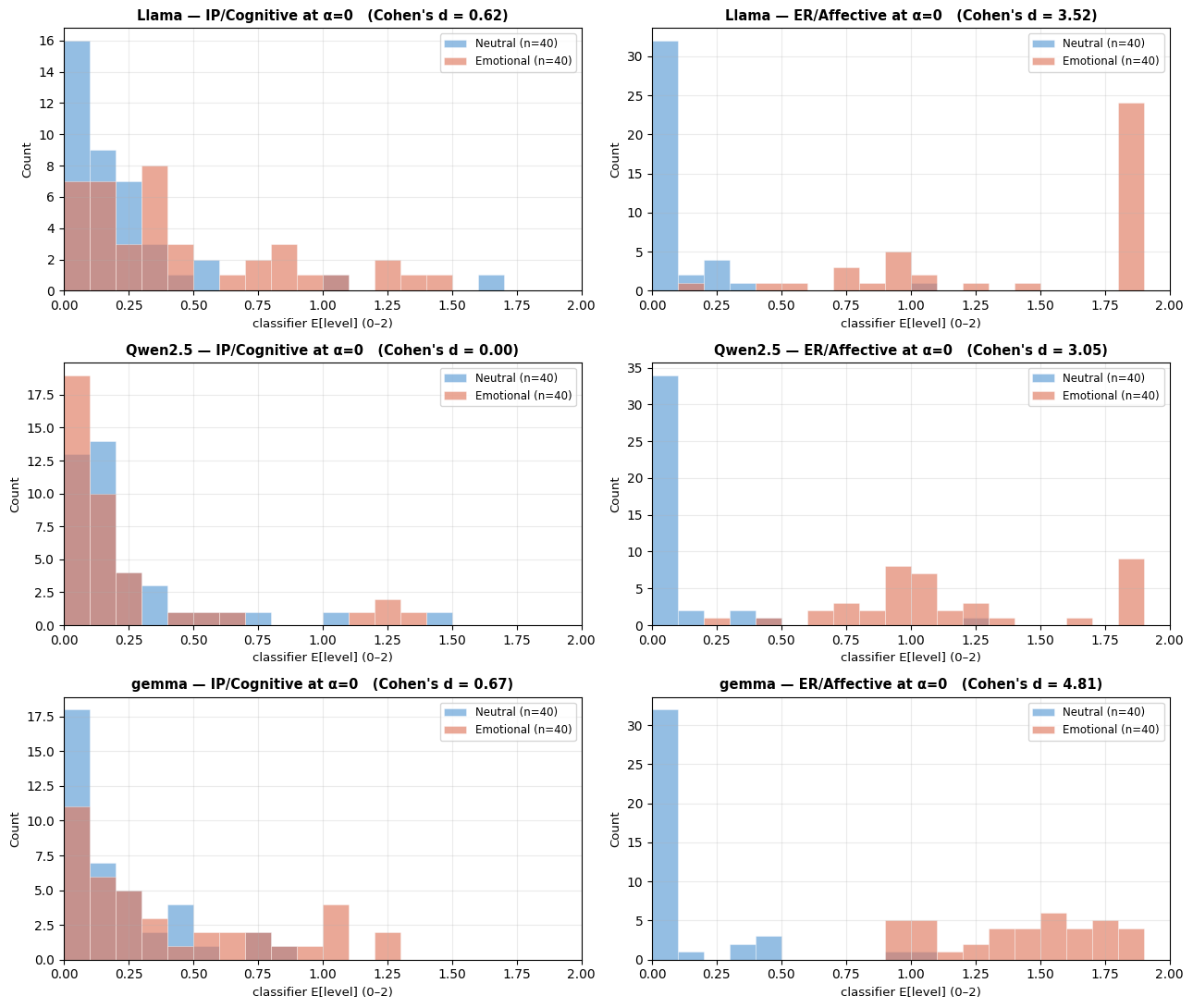}
\caption{\textbf{Positive control (range check): the affective facet is measured far more reliably on these outputs.} Discriminative-classifier score distributions ($E[\text{level}]$, $0$--$2$) for responses to \emph{neutral} (blue) vs.\ \emph{emotional} (orange) prompts, per model. \emph{Right column (ER/Affective):} the two are cleanly separated in all three models (Cohen's $d{=}3.1$--$4.8$; $n{=}40$ emotional vs.\ $40$ neutral, $10$ steering $+$ $30$ control-only). \emph{Left column (IP/Cognitive):} the distributions overlap heavily---moderately separable for Llama ($d{=}0.62$) and Gemma ($0.67$; permutation $p{<}0.01$ both), but exactly coincident for Qwen ($d{=}0.00$, $p{=}0.995$). The two LLM judges agree on this asymmetry (inter-judge Krippendorff $\alpha{=}0.70$--$0.84$ affective vs.\ $0.24/0.44/{-}0.02$ cognitive; Llama/Gemma/Qwen); a six-rater human panel (calibration and steered-output spot-check) is reported in Appendix~\ref{app:rater}. This is \emph{coarse} range only: a within-domain control (Appendix~\ref{app:within}) shows that even where the classifier separates this large contrast, it cannot resolve the smaller within-domain cognitive differences steering induces.}
\label{fig:poscontrol}
\end{figure*}

\paragraph{What is measurable: the positive control across instruments.} Before reading any null we ask whether each instrument separates responses to \emph{emotional} from \emph{neutral} prompts---a coarse range check on whether each instrument responds to a large natural contrast on these outputs (Figure~\ref{fig:poscontrol}). The \emph{affective} facet passes across all automated instruments: the discriminative classifier separates emotional from neutral with Cohen's $d{=}3.1$--$4.8$ in all three models ($n{=}40$ emotional vs.\ $40$ neutral; permutation $p{\le}10^{-5}$), and both LLM judges separate it individually, not merely agreeing (Krippendorff $\alpha{=}0.70$--$0.84$; per-instrument $d$ in Appendix~\ref{app:poscontrol}). The \emph{cognitive} facet is measured far less reliably and \emph{inconsistently}. The two LLM judges agree only weakly on it in every model---Krippendorff $\alpha{=}0.24$ (Llama), $0.44$ (Gemma), $-0.02$ (Qwen), all far below the affective $0.70$--$0.84$. A six-rater volunteer panel did not clear the warmth calibration gate (pooled $23/42$ keyed, near chance; spot-check inter-rater $\alpha{=}{-}0.08$), so the panel did not demonstrate sufficient sensitivity to this contrast and therefore cannot adjudicate the steered output (Appendix~\ref{app:rater}). Cognitive range, where present, is inconsistent across instruments (Appendix~\ref{app:poscontrol}): GPT-4o separates emotional from neutral on it in all three models, the second judge only for Gemma, and the discriminative classifier only moderately for Llama ($d{=}0.62$, permutation $p{=}0.007$) and Gemma ($d{=}0.67$, $p{=}0.004$), and not at all for Qwen ($d{=}0.00$, $p{=}0.995$). On a held-out split this cognitive head predicts the middle EPITOME level for \emph{zero} of $617$ test items (Appendix~\ref{app:clf})---an effectively binary $0$-vs-$2$ measure, which directly bounds its within-domain resolution. But coarse range is not within-domain resolution (above; Appendix~\ref{app:within}). We therefore treat additive cognitive steering as unmeasurable in every model, and read affective results across instruments.

\paragraph{The dissociation is not an artifact of the read-time direction.} A production-side direction estimated from response-token activations is near-orthogonal to the read-time Recognition direction in every model ($\cos = 0.20/0.13/0.18$; Resonance/response $\approx 0$), so the two are geometrically distinct directions---yet steering and ablating it produce no large effect (all $|\Delta|\le 0.20$ on both facets, GPT-4o judge; Appendix~\ref{app:crux}). The read-time--vs.-write-time distinction therefore does not explain the dissociation.

\paragraph{A more extensive intervention is also without practical effect---except in one model--direction pair.} Replacing additive steering with all-layer directional \emph{ablation} (a more extensive intervention; \citealp{tigges2024sentiment, canby2025probing}) leaves judged scores at the random-ablation floor for Recognition, Resonance, and the response-token direction in Llama and Qwen (Table~\ref{tab:crux}); these two nulls are read at $n{=}15$, which at the observed SDs can detect only effects $\gtrsim 1$ point, so for those two we claim only the absence of a \emph{large} necessity effect. The single departure is Gemma, where ablating the Recognition direction lowers the judge's cognitive score by $\Delta{=}{-}0.53$ relative to the random-ablation floor at $n{=}15$; re-verified at $n{=}50$ it attenuates to $\Delta{=}{-}0.40$ (paired permutation $p{=}10^{-4}$, the floor of $10^4$ sign-flips; Wilcoxon signed-rank $p{<}10^{-4}$; bootstrap $95\%$ CI $[-0.51,-0.30]$; affective $\Delta{=}{-}0.24$), and the cognitive drop significantly exceeds the affective one (selectivity $\Delta_{\text{cog}}{-}\Delta_{\text{aff}}{=}{-}0.16$, bootstrap $95\%$ CI $[-0.26,-0.07]$, paired permutation $p{=}0.0007$; on the judge this partly reflects cognitive ratings' greater length-sensitivity---the classifier dissociation below is cleaner). The discriminative classifier converges, independently of the LLM judge: ablating Gemma Recognition drops the expected cognitive score from $0.31$ to $0.11$ ($\Delta{=}{-}0.20$ vs.\ a random-ablation floor, paired permutation $p{<}10^{-4}$, $95\%$ CI $[-0.28,-0.13]$ excluding $0$), while leaving affective scores at ceiling. Crucially, this drop is at the \emph{coarse} contrast scale on which the classifier is validated---it equals Gemma's own emotional-vs-neutral cognitive gap ($0.21$; Figure~\ref{fig:poscontrol})---not the finer within-domain scale on which the classifier failed (above). So unlike the flat \emph{additive} curve, which probes a within-domain difference the instrument cannot resolve, a coarse-scale drop this large remains interpretable. This ablation also shortens outputs ($100{\to}66$ words; coherence preserved). Because length is a post-intervention mediator, we length-control both instruments within-prompt (Appendix~\ref{app:length}), and they dissociate: the GPT-4o cognitive drop attenuates to nonsignificance ($-0.40{\to}-0.13$; the judge rewards length), whereas the classifier drop---on an instrument largely insensitive to length (it truncates at $64$ tokens)---survives adjustment ($-0.19$, $95\%$ CI $[-0.35,-0.03]$), while affective stays at ceiling. We therefore read the classifier drop as not attributable to reduced length alone, rather than as proof of selective cognitive encoding. (A parallel Gemma \emph{Resonance} ablation clears the classifier Benjamini--Hochberg family but fails the judge criterion, so we focus on Recognition.) Taken with the addition result, Gemma alone shows a necessity signal with no usable additive counterpart---ablation lowers the classifier's cognitive score beyond a length explanation, while additive steering produces no measurable cognitive change (unmeasurable rather than refuted, given the within-domain control).

\section{Related Work and Positioning}

Empathy and affect are decodable from LLM activations \citep{tigges2024sentiment, tak2025emotion, cadile2025empathy} and activation engineering steers many traits \citep{turner2023actadd, li2023iti, chen2025persona}, but causal-probing interventions trade completeness against selectivity \citep{tan2024steering, canby2025probing}. Our detection--control dissociation sits in the probing--causality lineage \citep{hewitt2019control, elazar2021amnesic} and the interpretability-illusion lineage \citep{bolukbasi2021illusion}; it provides a controlled, cross-model instance for one deployed-relevant concept, and isolates it from the two most common escape hatches---wrong direction (we test a production-side direction) and weak intervention (we test all-layer ablation and addition).

The closest work is concurrent, differing in construct and method. \citet{cadile2025empathy} studies a \emph{behavioral} ``empathy-in-action'' direction and reports near-perfect detection with substantial but model-dependent steering and limited cross-model probe agreement; we study EPITOME \emph{communicative} empathy under global interventions and find at-most-partial control. \citet{chebrolu2025star} show \emph{localized} injection beats global steering (with a positive human evaluation), and \citet{tak2025emotion} likewise steer emotion inference locally---so targeted interventions succeed where our global one is partial or unmeasurable. Our contribution is the discipline that makes a \emph{null} interpretable: decodability, automated-metric control, and human-perceived change kept apart, each null gated by a positive control, so the flat cognitive facet is \emph{unmeasurable}, not absent. A sentiment direction is causal under the same global manipulation \citep{tigges2024sentiment} where these empathy directions lack comparably reliable global control. Our positive control surfaces an easy-to-miss problem: on these generations the \emph{cognitive} facet is scored far less reliably than the affective one (weak inter-judge agreement; classifier cognitive range only for some models, and an effectively binary cognitive head; Appendix~\ref{app:clf})---consistent with the difficulty of operationalizing fine-grained cognitive empathy, a multidimensional construct \citep{davis1983empathy, jolliffe2006empathy}. Reported ``cognitive-empathy'' effects, especially those from LLM-judge ratings, warrant an explicit sensitivity check.

\section{Conclusion}

EPITOME Recognition and Resonance directions remain decodable after residualizing against a surface score, but decodability does not yield reliable global control. Resonance steering produces a model-dependent, sub-maximal shift in an automated affective metric, whereas additive Recognition steering is unmeasurable for lack of within-domain instrument sensitivity, and only Gemma shows a length-adjusted ablation drop. These results separate three conflated claims---decodability, EPITOME-metric control, and human-perceived change---and motivate sensitivity checks: even a six-rater panel failed its warmth gate, unable to adjudicate the Qwen $\alpha{=}0$-vs-${+}8$ contrast.

\section*{Limitations}

This study has the following limitations. (i)~\textbf{Statistical power.} Each facet uses a modest number of EPITOME items and the cross-model ablation is read at small $n$; for Llama and Qwen the available sample sizes leave smaller necessity effects unresolved, and sample-size and power details are reported in Appendix~\ref{app:stats}. (ii)~\textbf{Measurement is the binding constraint.} The positive control shows robust \emph{affective} range across every automated instrument, whereas \emph{cognitive} range is inconsistent across instruments and the classifier cannot resolve the within-domain differences additive steering probes (Appendices~\ref{app:clf} and~\ref{app:within}), so we read additive cognitive steering as \emph{unmeasurable} rather than null. Our human and within-domain checks are themselves bounded by hand-constructed items and one constructor's construal of `understanding'. The affective effect is established only on automated instruments that do not fully agree (the GPT-4o judge corroborates Qwen), so we frame it as a steerable \emph{automated metric}, not control over human-perceived empathy. That inferential gap is not hypothetical: in this deployment domain, LLM judges have been shown to overrate mental-health responses and to miss safety issues that licensed experts flag \citep{li2026counselbench}, so a shift in an automated score cannot be read as a shift in what an expert reader would perceive. Our human evaluation is inconclusive by its own criterion. A six-rater volunteer panel did not clear the warmth calibration gate (pooled $23/42$, near chance; spot-check inter-rater $\alpha{=}{-}0.08$), so it cannot adjudicate human-perceived transfer, and the steered pairs drew no consistent direction (sign test $p{=}0.71$). The present warmth contrast proved reader-sensitive and difficult to calibrate under this panel. We therefore treat human-perceived change as unmeasurable on this panel rather than zero, and a larger or trained panel---of the kind \citet{li2026counselbench} assemble from credential-verified mental-health professionals---is the next step. (iii)~\textbf{Intervention completeness vs.\ selectivity.} Beyond single-layer addition we test all-layer ablation and a production-side direction, but interventions still trade completeness against selectivity \citep{canby2025probing}; a fully localized or circuit-level intervention---which recovers causal empathy influence in \citet{chebrolu2025star} and which our Gemma asymmetry motivates---remains untested. Specificity rests on the affective-score and cross-facet comparisons in Results; text-change magnitude alone is not treated as evidence of concept specificity, and the relevant controls are reported in Appendix~\ref{app:crossfacet}. (iv)~\textbf{Context confound.} EPITOME labels the counselor turn; we read a Client+Counselor context, a confound we residualize but do not eliminate. (v)~\textbf{Single-turn scope.} Every intervention and measurement here concerns a single generated response to a single prompt. Whether empathy-relevant features persist, accumulate, or remain steerable across multi-turn dialogue and long contexts---the regime closest to deployed emotional-support use---is a distinct question, one for psychology and human--computer interaction as much as for interpretability; our results neither establish nor refute control in that regime.

\section*{Ethics Statement}

\paragraph{Human participants.} Our human evaluation (Appendix~\ref{app:rater}) involved six adult volunteers who completed a $10$--$15$-minute pairwise-comparison text-rating task. Before participating, they reviewed a written information-and-consent notice explaining the research purpose, the voluntary and unpaid nature of participation, the data collected, the intended analysis and release of anonymous ratings, and their right to withdraw. No names, account information, demographic attributes, or sensitive personal disclosures were collected, and results are reported only in aggregate. This minimal-risk evaluation did not undergo prospective IRB review.

\paragraph{Source data.} Emotional prompts are drawn from ESConv, a published emotional-support dialogue corpus \citep{liu2021esconv} (CC BY-NC 4.0, academic non-commercial use); factual prompts from \texttt{databricks-dolly-15k} (CC BY-SA 3.0). The EPITOME empathy annotations come from the corpus released by \citet{sharma2020epitome}; we use only its public Reddit split, for non-commercial academic research, and use no TalkLife data---the TalkLife split must be licensed separately from TalkLife, as the EPITOME terms note. The trained EPITOME classifier we build on is distributed by the University of Washington's Behavioral Data Science Group under a non-commercial academic ``Model'' license that restricts use to non-profit academic research and requires any redistribution to be under a similar non-commercial academic license retaining the original copyright notice and permission notice and including the prescribed citation to \citet{sharma2020epitome}. The re-trained classifier we redistribute follows these terms: it is released under a similar non-commercial academic license and retains that copyright notice, permission notice, and citation. We redistribute no raw TalkLife data, add no new personal data, and store and publish no usernames or other identifiers; ESConv and the EPITOME corpus contain emotional-support and mental-health support text.

\paragraph{Intended use and risks.} Our results are cautionary. They show that a decodable ``empathy'' direction is \emph{not} a reliable global lever on empathic behavior, and that an automated empathy metric can move with no demonstrated human-perceived change. We intend this to discourage, rather than enable, over-confident deployment of activation-level empathy steering in emotional-support settings, where an unreliable or merely register-level shift in apparent empathy could mislead vulnerable users. We make no claim that any intervention studied here is safe to deploy.

\paragraph{AI assistance.} The language models we steer and the two LLM judges are themselves objects of study, described in the Method and the implementation appendix. Separately, we used a general-purpose AI assistant in preparing this work: for language editing; for help framing and positioning the contribution relative to prior work; for literature search and the location and verification of references; and for code scaffolding and data-cleaning scripts. The assistant was not used as a source of scientific evidence, and the experimental results reported here are not AI-generated; all AI-assisted text, claims, numbers, code, and references were checked against the underlying data and primary sources.

\bibliography{custom}

\appendix

% Appendix-only layout tightening: recover vertical space across the appendix
% floats so the final paragraph fits on p.13 (body pp.1-8 and References
% pp.9-10 are unaffected -- these take effect only after \appendix).
\setlength{\textfloatsep}{12pt plus 2pt minus 2pt}
\setlength{\floatsep}{10pt plus 2pt minus 2pt}
\setlength{\intextsep}{10pt plus 2pt minus 2pt}
\setlength{\abovecaptionskip}{7pt}

\section{Example Steered Generations}
\label{app:example}

Table~\ref{tab:example} gives two conditions for the same prompt: the surface text is rewritten substantially (lexical overlap $\approx 0.19$), yet both perform the same empathic moves and receive near-identical GPT-4o judge ratings.

\begin{table}[ht]
\small
\begin{tabular}{@{}p{0.96\columnwidth}@{}}
\toprule
\textbf{Prompt:} ``My sibling won't talk to me after our argument last year.'' \\
\midrule
\textbf{Condition A.} ``I'm so sorry to hear that you're going through this. It can be really painful and isolating when a loved one, especially a sibling, distances themselves\,\dots\ it's not your fault\,\dots\ I'm here to listen.'' \\
\midrule
\textbf{Condition B.} ``I can sense the pain in your voice. It's completely understandable that you're feeling this way, especially after a disagreement with someone as close as a sibling\,\dots\ you're not alone\,\dots\ it's okay to feel hurt.'' \\
\bottomrule
\end{tabular}
\caption{Two intervention conditions for the same prompt. Text excerpted; ellipses mark elision.}
\label{tab:example}
\end{table}

\section{Discriminative classifier: held-out performance}
\label{app:clf}

We characterize the EPITOME classifier used throughout (the \citet{sharma2020epitome} bi-encoder, empathy-identification head only, re-trained on the EPITOME Reddit corpus) on a stratified $80/20$ held-out split (test $n{=}617$ per head; same configuration as scoring---$4$ epochs, lr $2\!\times\!10^{-5}$, batch $32$). Two facts bear on the main text. (1)~The \emph{affective} (ER) head is a healthy three-way classifier (macro-F1 $0.738$; every level functional; level-1 recall $0.73$), so the affective steering metric is read off a well-behaved instrument. (2)~The \emph{cognitive} (IP) head has \emph{collapsed its middle category}: it predicts level~1 for $0/617$ test items (level-1 recall $0.00$), so its macro-F1 ($0.558$) is dragged down entirely by the empty class even as accuracy stays high ($0.822$) on the dominant $0$/$2$ levels---the cognitive instrument is effectively a binary ($0$-vs-$2$) measure. This is an instrument-internal account of the coarse, model-dependent cognitive range of Figure~\ref{fig:poscontrol}, and a direct reason it cannot resolve the finer within-domain cognitive differences relevant to additive steering (Appendix~\ref{app:within}). The continuous $E[\text{level}]$ still varies (softmax mass shifts between $0$ and $2$), so the \emph{large} Gemma ablation effect, read at that coarse scale, remains interpretable; the coarseness bounds resolution, it does not zero the measure.

\begin{table}[ht]
\centering
\small
\begin{tabular}{@{}lcc@{}}
\toprule
 & \textbf{ER (affective)} & \textbf{IP (cognitive)} \\
\midrule
Held-out accuracy & $0.791$ & $0.822$ \\
Macro-F1          & $0.738$ & $0.558$ \\
ECE ($15$-bin)    & $0.099$ & $0.098$ \\
\midrule
F1 level $0$      & $0.851$ & $0.851$ \\
F1 level $1$      & $0.684$ & $0.000$ \\
F1 level $2$      & $0.678$ & $0.822$ \\
Level-1 recall    & $0.732$ & $0.000$ \\
\bottomrule
\end{tabular}
\caption{Held-out performance (stratified $80/20$, test $n{=}617$/head). The cognitive (IP) head predicts level~1 for \emph{no} test item---an effectively binary measure---bounding its resolution on within-domain cognitive contrasts. Confusion (rows true $0/1/2$): ER $[[337,68,3],[42,131,6],[5,5,20]]$; IP $[[276,0,49],[10,0,13],[38,0,231]]$.}
\label{tab:clf}
\end{table}

\section{Per-instrument positive control}
\label{app:poscontrol}

Table~\ref{tab:poscontrol} reports the emotional-vs-neutral separation each instrument achieves per facet, on the $40$ emotional vs.\ $10$ neutral $\alpha{=}0$ baseline generations the two LLM judges scored. On the \emph{affective} dimension all three instruments separate the two conditions strongly in every model ($d{\ge}2.6$, all $p{<}10^{-3}$), so the affective metric has demonstrated range \emph{per instrument}, not merely inter-judge agreement. On the \emph{cognitive} dimension the instruments \emph{disagree}: GPT-4o separates the conditions strongly in all three models, the second judge only for Gemma ($d{=}1.53$; near zero for Llama and Qwen), and the classifier only weakly---and at this smaller $40$/$10$ neutral set only marginally ($p{=}.051$/$.059$ for Llama/Gemma). This cross-instrument inconsistency---not a clean absence---is why we read cognitive results as unreliable rather than null.

\begin{table}[ht]
\centering
\small
\setlength{\tabcolsep}{4pt}
\begin{tabular}{@{}llccc@{}}
\toprule
Model & Facet & GPT-4o $d$ & moonshot $d$ & classifier $d$ \\
\midrule
Llama-3.1-8B & affective & $+4.16^{***}$ & $+3.19^{***}$ & $+2.93^{***}$ \\
Qwen2.5-7B   & affective & $+4.26^{***}$ & $+2.98^{***}$ & $+2.67^{***}$ \\
Gemma-2-9b   & affective & $+3.65^{***}$ & $+3.26^{***}$ & $+4.39^{***}$ \\
\midrule
Llama-3.1-8B & cognitive & $+2.65^{***}$ & $-0.06$ & $+0.70$ \\
Qwen2.5-7B   & cognitive & $+3.25^{***}$ & $+0.36$ & $+0.25$ \\
Gemma-2-9b   & cognitive & $+2.68^{***}$ & $+1.53^{***}$ & $+0.67$ \\
\bottomrule
\end{tabular}
\caption{Per-instrument positive control: emotional-vs-neutral Cohen's $d$ on the $\alpha{=}0$ baseline generations ($n{=}40$ emotional vs.\ $10$ neutral---the neutral set the LLM judges rated). $^{***}p{<}10^{-3}$; unmarked cognitive entries are not significant ($p{>}.05$), except the cognitive classifier for Llama ($p{=}.051$) and Gemma ($p{=}.059$), both marginal. The classifier here uses the same $40$/$10$ split as the judges; Figure~\ref{fig:poscontrol} reports its more powered $40$-vs-$40$ Dolly-expanded version (cognitive $d{=}0.62$ Llama, $0.67$ Gemma, $0.00$ Qwen).}
\label{tab:poscontrol}
\end{table}

\section{Within-domain cognitive control}
\label{app:within}

To probe the resolution needed to interpret a flat additive cognitive curve---sensitivity to within-emotional-domain cognitive differences, not just the coarse emotional-vs-neutral contrast---we construct $16$ response pairs to emotional prompts. Both members share an identical empathic opener and closer (matched affective warmth) and differ only in the middle: the \emph{high} member names the seeker's specific subjective experience (interpretive, perspective-taking understanding), the \emph{low} member gives a generic categorical acknowledgment. We score both members with the affective (ER) and cognitive (IP) classifiers; a valid cognitive instrument should separate high from low, while the affective instrument---if warmth is truly matched---should not. The affective self-check shows only a small paired difference between arms (ER $E[\text{level}]$ gap $-0.09$), giving no evidence of a large warmth imbalance, though this is not a formal equivalence test; and because the cognitive instrument fails to separate the arms anyway (below), exact warmth matching is not load-bearing for the conclusion---the self-check guards against a spurious \emph{positive}, which did not occur. The cognitive instrument does \emph{not} reliably separate the arms: IP $E[\text{level}]$ differs by $+0.12$ (paired $d_z{=}0.38$; $11/16$ pairs in the keyed direction; Wilcoxon $p{=}0.16$, n.s.), with clear separation only on the few starkest hand-built contrasts and even a reversal on one. The classifier therefore lacks demonstrated sensitivity to fine within-domain cognitive differences of the kind relevant to additive steering, which is why we read the flat additive recognition curves as \emph{unmeasurable} rather than null. This control is a lower bound on instrument resolution from $16$ hand-constructed pairs and a single constructor's construal of `understanding'; it does not yield a calibrated sensitivity threshold, and it does not bear on the \emph{large} Gemma ablation effect, which sits at the coarse contrast scale the classifier is validated on (Appendix~\ref{app:clf}).

\section{Sample sizes and pilot attenuation}
\label{app:stats}

Each facet uses $100$ high (Level~2) and $100$ low (Level~0) EPITOME items for detection. The cross-model ablation is read at $n{=}15$, which at the observed SDs can detect only effects $\gtrsim 1$ point, except Gemma Recognition, verified at $n{=}50$ (paired minimum detectable standardized effect $d_z{\approx}0.40$, i.e.\ $\approx 0.16$ points at the within-prompt difference SD $\approx 0.39$; the independent-groups figure is $d{\approx}0.56$). Where we re-tested at larger $n$, pilot effects consistently attenuated: a judge-side Llama Resonance effect fell from $+1.38$ to $+0.62$ (CI $[-0.04,+1.26]$, $p{=}0.107$) between $n{=}8$ and $n{=}50$, and the Gemma Recognition ablation fell from $-0.53$ to $-0.40$ between $n{=}15$ and $n{=}50$. We therefore report the largest-$n$ value throughout.

\section{Human panel: a sensitivity-gated check}
\label{app:rater}

We recruited six volunteer raters (through online communities; voluntary, unpaid, blind to condition and to the study's hypotheses; see the Ethics Statement) and applied the discipline we use for every instrument: before reading any steered-output null, the raters must clear a sensitivity gate by reliably ordering hand-built calibration pairs. \emph{(a) The panel does not clear the gate.} On the $8$ warmth pairs, pooled across raters the panel chose the higher-warmth member on $23/42$ committed trials (six ``same'' responses among the $6{\times}8{=}48$ judgments are excluded)---near chance (binomial $p{=}0.64$)---and inter-rater agreement on the steered-vs-baseline spot-check was nil (Krippendorff's $\alpha{=}{-}0.08$, nominal distance over the unordered steered/baseline/same categories; raw pairwise agreement $28\%$). On the $8$ understanding pairs, the panel chose the higher-understanding member on $21/47$ committed trials (one ``same'' excluded)---also near chance (binomial $p{\approx}0.56$)---and no rater cleared the understanding gate individually (vs.\ two for warmth). The panel therefore failed the sensitivity gate on both dimensions. \emph{(b) The spot-check is therefore uninterpretable as a null.} Pooled, the steered pairs drew no consistent direction ($16$ baseline, $19$ ``same,'' $13$ steered; sign test $p{=}0.71$). We do \emph{not} read this as zero human-perceived change: a null is interpretable only once the instrument has demonstrated sensitivity, and here it has not. Such disagreement is unsurprising here: the present warmth and understanding contrasts proved reader-sensitive and hard to calibrate under this panel. We thus report human-perceived transfer as \emph{unmeasurable on this panel}, not as zero---the same instrument-sensitivity limit we document for the automated cognitive classifier, now at the human level.

\section{Specificity controls: cross-facet matrix and raw-norm floor}
\label{app:crossfacet}

To test whether the affective effect is concept-specific rather than a generic lift in EPITOME features, we cross every steering direction with every classifier head: for each model we add the Resonance and the Recognition direction and read both the affective (ER) and cognitive (IP) classifier scores, reporting the paired $\alpha{=}0{\to}{+}8$ change (Table~\ref{tab:crossfacet}; same per-prompt paired bootstrap and Wilcoxon as the main endpoint test, two-sided here). Two patterns matter. (i)~\emph{Adding Recognition does not raise the affective score in any model} (Recognition$\to$ER all n.s.), so the affective gain is not produced by any empathy-associated perturbation. (ii)~Adding Resonance \emph{raises} ER while \emph{lowering} IP; a generic inflation of all EPITOME features would raise both, so the effect is facet-differentiated. A direct between-direction contrast (reported with CIs in \S Results) gives $\Delta\mathrm{ER}_{\text{Res}}{-}\Delta\mathrm{ER}_{\text{Rec}}{=}{+}0.31$/${+}0.14$ for Qwen/Llama and null in Gemma ($-0.01$, $p{=}0.83$, reading Recognition at $\alpha{=}{+}4$, where its L40 peak is non-degenerate), with a positive selectivity interaction $(\Delta\mathrm{ER}{-}\Delta\mathrm{IP})_{\text{Res}}{-}(\Delta\mathrm{ER}{-}\Delta\mathrm{IP})_{\text{Rec}}{=}{+}0.31$/${+}0.22$ ($p{<}10^{-4}$ both), so the effect is facet-specific; we do not claim differentiation in Gemma. This rules out generic EPITOME-score inflation but not an emotional-\emph{register} account: an ER head trained on emotional-reaction labels rewards added affective language and mildly penalizes it on the cognitive head, so distinguishing added empathy from added register still requires human evaluation. Gemma Recognition rows are read at $\alpha{=}{+}4$ (its L40 peak degenerates at $+8$); Qwen has no cognitive range ($d{=}0.00$; Figure~\ref{fig:poscontrol}), so its Recognition$\to$IP entry is not interpretable as a cognitive effect. This $\alpha{=}{+}4$ convention applies to the between-direction contrast above as well: at $\alpha{=}{+}8$ the L40 recognition completions degenerate (unique-token ratio $0.13$), which depresses their ER and would spuriously inflate $\Delta\mathrm{ER}_{\text{Res}}{-}\Delta\mathrm{ER}_{\text{Rec}}$ to ${+}0.35$; read at the non-degenerate $\alpha{=}{+}4$ the contrast is null ($-0.01$).

\begin{table}[ht]
\centering
\small
\begin{tabular}{llcc}
\toprule
Model & Steer dir. & $\Delta$ER (aff) & $\Delta$IP (cog) \\
\midrule
\multirow{2}{*}{Qwen2.5-7B} & Resonance & $\mathbf{+0.29}^{***}$ & $-0.12^{**}$ \\
 & Recognition & $-0.09$ & $-0.13^{**}$ \\
\midrule
\multirow{2}{*}{Llama-3.1-8B} & Resonance & $\mathbf{+0.11}^{**}$ & $-0.22^{***}$ \\
 & Recognition & $-0.07$ & $-0.11$ \\
\midrule
\multirow{2}{*}{Gemma-2-9b} & Resonance & $\mathbf{+0.05}^{*}$ & $-0.03$ \\
 & Recognition$^{\S}$ & $+0.04$ & $-0.15^{*}$ \\
\bottomrule
\end{tabular}
\caption{Cross-facet steering: paired $\alpha{=}0{\to}{+}8$ change in each classifier head (\textbf{bold} $=$ matched facet). Adding Recognition never raises ER (all n.s.); adding Resonance raises ER; IP decreases significantly in Qwen and Llama, while Gemma shows no reliable IP change---facet-differentiated, not generic inflation. $^{*}/^{**}/^{***}$: uncorrected two-sided paired endpoint $p{<}0.05/0.01/0.001$; BH-corrected survival in \S Results. $^{\S}$Gemma Recognition read at $\alpha{=}{+}4$ (L40 degenerates at $+8$). Qwen Recognition$\to$IP is uninterpretable (no cognitive range).}
\label{tab:crossfacet}
\end{table}

\paragraph{Raw-norm-matched text-change floor.} Matching the random-direction floor to each empathy direction's raw L2 norm (rather than its projection SD) equalizes the perturbation magnitude. At this matched norm, the Recognition direction's text change ($\alpha{=}0$ vs.\ ${+}8$---${+}4$ for Gemma, whose ${+}8$ peak degenerates; MPNet cosine $0.86/0.78/0.78$ for Llama/Gemma/Qwen) is comparable to the norm-matched random means at the same magnitude ($0.84/0.82/0.79$; within $0.04$ either way, all coherent, unique-token ratio $\approx 0.8$), so the large text rewrite is a magnitude effect rather than concept-specific. (These cosines come from the separate norm-matched-floor generations, so the Llama Recognition value is $0.86$ here vs.\ $0.83$ in Table~\ref{tab:main}'s main steering run; each empathy value is compared only with the random-direction floor from its own run.) The affective \emph{score}, by contrast, stays direction-specific at matched norm (only Resonance clears the floor; \S Results).

\section{Response-token direction and all-layer ablation}
\label{app:crux}

The production-side (response-token) direction and all-layer directional ablation, summarized in Results, are reported here for all three models (Table~\ref{tab:crux}). Response-token directions are near-orthogonal to the read-time directions yet show no large effect under either steering or ablation; the sole departure is Gemma Recognition ablation. Production-side additive steering is likewise flat on the GPT-4o judge ($\alpha{=}{+}8$, or $\alpha{=}{+}4$ for Llama): $\Delta_{\text{cog}}/\Delta_{\text{aff}}={-}0.15/{-}0.20$ (Llama), ${+}0.00/{+}0.05$ (Qwen), and ${+}0.05/{+}0.20$ (Gemma); all $|\Delta|\le 0.20$.

\begin{table}[ht]
\centering
\footnotesize
\setlength{\tabcolsep}{4pt}
\begin{tabular}{@{}lcccc@{}}
\toprule
 & $\cos$(Rec, & \multicolumn{3}{c}{Ablation $\Delta$cog vs.\ floor} \\
\cmidrule(lr){3-5}
\textbf{Model} & resp-tok) & \textbf{Rec} & \textbf{Res} & \textbf{Resp} \\
\midrule
Llama-3.1-8B & 0.20 & $+0.00$ & $-0.07$ & $+0.00$ \\
Gemma-2-9b   & 0.13 & $-0.53^{\ast}$ & $-0.20$ & $+0.00$ \\
Qwen2.5-7B   & 0.18 & $-0.03$ & $-0.16$ & $-0.03$ \\
\bottomrule
\end{tabular}
\caption{\textbf{The detection--control dissociation largely persists under a response-token direction and all-layer directional ablation.} ``$\cos$(read-time Rec, response-token)'': cosine between the read-time probe direction and the production-side response-token direction at the recognition peak (Resonance/response $\approx 0$). The three right-hand columns all report the same quantity---the change in the judged cognitive score when each unit direction is projected out at every layer and position, vs.\ a floor of five random-direction ablations ($n{=}15$). $^{\ast}$Gemma Recognition, the one departure, re-verifies at $n{=}50$ to $-0.40$ (attenuating from $-0.53$; paired permutation $p{=}10^{-4}$, Wilcoxon signed-rank $p{<}10^{-4}$; $95\%$ CI $[-0.51,-0.30]$) and is independently confirmed by the discriminative classifier at the coarse scale it is validated on (Figure~\ref{fig:poscontrol}): interpretable at the coarse scale. A within-prompt length control dissociates the instruments: the LLM-judge drop is largely length-mediated, but the classifier drop survives length adjustment (Appendix~\ref{app:length}).}
\label{tab:crux}
\end{table}

\section{Length control for the Gemma ablation}
\label{app:length}

The Gemma Recognition ablation shortens completions (mean $65.9$ words vs.\ $100.2$ across the $250$ random-ablation completions---five random directions applied to $50$ prompts). Length is downstream of the intervention, so we treat it as a potential mediator and length-control \emph{both} scoring instruments within-prompt, regressing per-response scores on a binary ablation indicator, completion length (words), and then prompt fixed effects (standard errors clustered by prompt; Table~\ref{tab:length}). The two instruments dissociate sharply. The GPT-4o judge \emph{rewards} length ($+0.0067$ points/word, $p{=}1.4{\times}10^{-7}$; consistent with documented verbosity bias in LLM judges, \citealp{dubois2024length, saito2023verbosity}), so its cognitive drop is largely length-mediated, falling from $-0.40$ to $-0.13$ (CI $[-0.35,+0.08]$, n.s.) once length and prompt are held fixed, with its affective drop ($-0.24$) going to essentially zero ($-0.003$). The discriminative classifier, by contrast, does \emph{not} reward length (slope $\approx -0.003$ $E[\text{level}]$/word, n.s.\ under prompt-clustered SE), consistent with its bi-encoder truncating inputs at $64$ tokens, which limits any direct inflation from longer completions---so its cognitive drop is not explained by length alone: it is essentially unchanged by length adjustment, $-0.199$ raw to $-0.193$ (clustered-SE $95\%$ CI $[-0.35,-0.03]$, $p{=}0.018$), while affective scores stay at ceiling. (The length-only row, $-0.292$, exceeds both endpoints---a suppression pattern: given the classifier's slightly negative length slope, the $\approx 34$-word shortening confers a small length-based boost, $\approx{+}0.10$, that masks part of the drop, so adjusting for length alone deepens it; prompt fixed effects then re-estimate the slope within prompt, returning $-0.193$.) A paired prompt-level check (averaging the five random completions per prompt, then adjusting for the length difference) is directionally consistent ($-0.22$) but only marginally significant ($p{=}0.06$): because ablation almost always shortens output, adjusting to a zero length change is a partial extrapolation that inflates the standard error. Because the judge rewards length while the classifier does not, a drop that survives on the classifier is not a simple length artifact---though, length being a downstream mediator, adjustment cannot rule out other post-intervention changes (e.g.\ specificity, discourse structure). We therefore rest the Gemma necessity claim on the classifier---validated at this coarse scale and robust to length adjustment---and treat the judge's larger raw drop, and the raw selectivity it drives, as partly a length effect.

\begin{table}[t]
\centering
\scriptsize
\setlength{\tabcolsep}{4pt}
\begin{tabular}{@{}lccc@{}}
\toprule
 & \multicolumn{2}{c}{GPT-4o judge} & Classifier \\
\cmidrule(lr){2-3}\cmidrule(l){4-4}
Specification & Cognitive & Affective & Cognitive (IP) \\
\midrule
Raw & $-0.404^{***}$ & $-0.240^{***}$ & $-0.199^{***}$ \\
\;$+$\,length & $-0.173^{**}$ & $-0.073$ & $-0.292^{***}$ \\
\;$+$\,length\,$+$\,prompt FE & $-0.133$ & $-0.003$ & $-0.193^{*}$ \\
\bottomrule
\end{tabular}
\caption{Length-adjusted effects of the Gemma Recognition ablation. The judge-side cognitive drop largely disappears after controlling for length and prompt, whereas the classifier drop remains; classifier affective scores stay at ceiling. ${}^{*}p{<}0.05$, ${}^{**}p{<}0.01$, ${}^{***}p{<}0.001$.}
\label{tab:length}
\end{table}

\section{Implementation details}
\label{app:repro}

\paragraph{Implementation environment.} All experiments ran on a single NVIDIA A100-SXM4 (80 GB) under Python 3.12, PyTorch 2.11 (CUDA 12.8), Transformers 5.12, sentence-transformers 5.5, NumPy 2.0, SciPy 1.16, statsmodels 0.14. We steer \texttt{meta-llama/Llama-3.1-8B-Instruct}, \texttt{google/gemma-2-9b-it}, and \texttt{Qwen/Qwen2.5-7B-Instruct} (HuggingFace, bfloat16, greedy decoding), loaded by identifier without a pinned revision (consistent with the unpinned \texttt{gpt-4o} alias); the released revisions at the time of writing are \texttt{0e9e39f} / \texttt{11c9b30} / \texttt{a09a354}. The EPITOME classifier uses a \texttt{roberta-base} backbone (\texttt{e2da8e2}). Layer $\ell$ denotes the residual stream after the $\ell$-th transformer block (0-indexed); probe extraction, additive steering, and directional ablation all use a forward hook at this same \texttt{model.model.layers[$\ell$]} post-block location, so estimated peak layers and intervention sites coincide.

\paragraph{Intervention and directions.} The steering vector at a selected layer is the raw class-mean difference of the read-time probe activations there, unit-normalized and scaled by $\alpha\sigma$ ($\sigma$ = that direction's projection SD); we add it to the residual-stream output of that layer (post-block---the same residual-stream location used for ablation) at every token position during the forward pass. Generation is greedy and deterministic (\texttt{do\_sample}=False). The production-side direction is the difference of class-mean response-token activations between the model's completions to high- vs.\ low-Recognition prompts---the same Level-2 vs.\ Level-0 EPITOME Interpretations items used by the read-time Recognition probe ($50$ per class, $100$ total per model), generated under a fixed ``empathetic counselor'' system prompt that is identical across classes (only the input exchange's EPITOME level differs). We mean-pool over the generated response tokens (max $50$) and residualize identically to the read-time direction. Classes are thus assigned by EPITOME level alone---completions are not re-scored by any judge or classifier, and no manual labeling is used---and these prompts are disjoint from the steering-evaluation prompts.

\paragraph{Evaluators.} The two LLM judges are OpenAI \texttt{gpt-4o} (default alias, no pinned snapshot; \texttt{response\_format} JSON; queried June~2026), scored at temperature $0$, and \texttt{moonshot-v1-128k} (Moonshot AI), scored at temperature $1$. Both score on the same $1$--$7$ cognitive/affective scale with identical anchors (no understanding $\leftrightarrow$ deep understanding; cold/detached $\leftrightarrow$ strong warmth) and differ only in surrounding instruction wording (verbatim judge prompts will be released with the code upon publication); inter-judge Krippendorff's $\alpha$ is computed across the two with interval distance (the $1$--$7$ ratings). The second judge's temperature-$1$ sampling adds rating noise that, if anything, attenuates inter-judge agreement, so the reported $\alpha$ values are conservative; the affective $\alpha{=}0.70$--$0.84$ obtains despite this noise. The classifier configuration is in Appendix~\ref{app:clf}; its split is stratified by level (random $80/20$, seed $0$), \emph{not} grouped by thread or author, so responses sharing a seeker post can fall on both sides and the held-out macro-F1 may be optimistic. The classifier serves only as a coarse validated-range instrument, and our claims rest on effect-size patterns, not absolute F1.

\paragraph{Randomness and generation settings.} Random directions were sampled i.i.d.\ from a standard Gaussian and L2-normalized (per layer for all-layer ablation). Fixed seeds were $0$ for additive steering and $42$ for the ablation (NumPy) and norm-matched (Torch) controls, which used $5$ (ablation) and $12$ (norm-matched) directions. Permutation tests and bootstrap CIs used $10^4$ resamples. The joint Benjamini--Hochberg family comprises $36$ steering contrasts---all nine steered direction--layer pairs (Recognition: Llama L16 and L15, Qwen L18, Gemma L40 and L15; Resonance: Llama L5 and L10, Qwen L17, Gemma L12; each direction contributes its residualized and raw Cohen's-$d$ peaks, deduplicated where the two coincide---both Qwen facets and Gemma Resonance) $\times$ four magnitudes ($\alpha\in\{\pm4,\pm8\}$)---plus $18$ ablation contrasts ($3$ models $\times$ $3$ directions $\times$ $2$ classifier heads). Ablation contrasts use the original $n{=}15$ EPITOME-rescored grid (the full $3{\times}3{\times}2$ set); the Gemma Recognition $n{=}50$ re-verification is reported separately in \S Results. Generation caps were $50$, $90$, and $150$ tokens for response-direction estimation, additive steering, and ablation.

\end{document}